\documentclass[12pt,twoside]{article}
\usepackage[
  margin=2.26cm
]{geometry}
\usepackage{graphicx}
\usepackage{amsmath}
\usepackage{amssymb}
\usepackage[T1]{fontenc}
\usepackage[utf8]{inputenc}
\usepackage{crimson}
\usepackage[misc]{ifsym}

\newif\ifblind
\blindfalse  

\usepackage[hyphens]{url}

\usepackage[authordate,giveninits=false,backend=biber,sortcites=true]{biblatex-chicago}
\usepackage[hyperfootnotes=false]{hyperref}
\usepackage[shortcuts]{extdash}

\usepackage{float}

\usepackage{xcolor}

\usepackage[most]{tcolorbox}
\tcbset{
  abstractbox/.style={
    colback=gray!5,
    colframe=gray!5,
    boxrule=0.3pt,
    arc=1mm,
    left=6pt,
    right=6pt,
    top=4pt,
    bottom=4pt,
    boxsep=5pt
  }
}

\definecolor{refcolor}{rgb}{0.1, 0.1, 0.4}

\hypersetup{
    colorlinks=true,
    linkcolor=refcolor,
    citecolor=refcolor,
    urlcolor=refcolor
}

\usepackage[activate={true,nocompatibility},final,auto=true,tracking=true,kerning=true,expansion=true,spacing=true,factor=1000,stretch=20,shrink=20,letterspace=0]{microtype}

\usepackage[font=small,labelfont=bf]{caption}

\usepackage{fancyhdr}
\usepackage{lastpage}

\newcommand{\doctitle}{Why We Care About Understanding}

\ifblind
  \newcommand{\docauthors}{}
\else
  \newcommand{\docauthors}{Matthieu Queloz \& Pierre Beckmann}
\fi

\fancypagestyle{plain}{%
  \fancyhf{}
  
  \fancyhead[R]{September 2026}
  \fancyhead[L]{Draft Manuscript}
}

\usepackage{authblk}

\title{\textbf{Why We Care About Understanding:\\Competence through Predictive Compression}}

\ifblind
  \author{}
  \date{}
\else
  \author[1]{Matthieu Queloz}
  \author[2,3]{Pierre Beckmann}
  \affil[1]{University of Bern, Department of Philosophy}
  \affil[2]{École Polytechnique Fédérale de Lausanne (EPFL)}
  \affil[3]{Idiap Research Institute}
  \affil[ ]{\textcolor{gray}{\small\Letter} \hspace{0.05cm} \href{mailto:matthieu.queloz@unibe.ch}{\texttt{matthieu.queloz@unibe.ch}}}
\fi

\makeatletter
\renewcommand{\maketitle}{\bgroup\setlength{\parindent}{0pt}
\thispagestyle{plain}
\begin{flushleft}
\vspace*{0.7cm}
  {\fontsize{17}{17}\selectfont \@title}
  \vspace{0.5cm}

  \@author

\end{flushleft}\egroup
}
\makeatother

\usepackage{titlesec}

\titleformat{\section}
  {\fontsize{17}{15}\selectfont\bfseries}
  {\thesection}{1em}{}

\begin{document}

\maketitle
\vspace{0.2cm}

\begin{tcolorbox}[abstractbox]
\noindent
 \textbf{Abstract:}
What is the relation between understanding and compression, and why does human understanding take such a heavily compressed form? Across information theory, machine learning, and AI research, a substantial tradition identifies understanding with compression—a thought captured in Gregory Chaitin's dictum that ``comprehension is compression.'' Philosophers, by contrast, have characterized understanding in terms of grasping connections, giving explanations, and handling novelty. This paper bridges the two pictures through three interlocking theses. The first concerns the \emph{concept} of understanding: it serves as an efficient \emph{proxy} for a distinctive form of robust competence, enabling us to identify whom to trust and whom to learn from. The second concerns the \emph{state} of understanding: to understand a domain is to possess a mental model of its relational structure that enables prediction, and what enables prediction enables compression, because what becomes predictable need not be stored separately. Compression is therefore not identical with comprehension, but its representational shadow. The third concerns the \emph{characteristically human form} of understanding: the fiduciary and transmission functions highlighted by the first thesis impose pressures of demonstrability and transmissibility that drive human understanding toward principled simplicity. The resulting framework explains both the appeal and the limits of compressionist accounts of understanding while shedding light on the inscrutability of AI systems.

\vspace{0.3cm}
\par
\noindent
\textbf{Word count:} 10,400 words excl. notes.
\end{tcolorbox}
\vspace{0cm}
\section{Introduction}\label{introduction}

When Johannes Kepler distilled Tycho Brahe's records of planetary movements into the laws of planetary motion, this was not just an advance in our understanding of the heavens, but also an extraordinary feat of representational economy: he compressed two decades of observations into three compact laws. Here, as in many other cases, understanding went hand-in-hand with compression. This conjunction has been interpreted as indicating a deeper connection. Across information theory, machine learning, and AI research, a substantial body of work has come to see compression as the computational face of understanding \parencite{chaitin2002_intelligibility, chaitin2006_limits_of_reason, schmidhuber2006_developmental, schmidhuber2008_compression, schmidhuber2010, Maguire2015CompressionismAT, zenil2019, wolfram2018_logic, deletang2024language, yu2024white, Li2024UnderstandingIC, ramstead2025noumenallabswhitepaper, hutter2005_uai, zhang2025theory}---to the point of identifying the two, as in the mathematician Gregory Chaitin's dictum that ``understanding is compression, comprehension is compression!'' \parencite[35]{chaitin2006_limits_of_reason}.

The question of how compression and understanding are related has acquired new urgency with the rise of large language models.\footnote{For representative positions across the contemporary spectrum, see \textcite{chollet2019measure}, \textcite{schmidhuber2010}, and \textcite{mitchellkrakauer2023}. For recent empirical investigations of the compression-comprehension link in large models, see \textcite{shani2025tokens}, \textcite{kumar2025questioning}, and \textcite{li2025tracing}; for the question's broader implications for cognitive science, see \textcite{serre2025prediction}. For representative public exchanges, see Chollet's threads at \url{https://x.com/fchollet/status/1727855160683372969} and \url{https://x.com/fchollet/status/1985753756873736304} (an exchange involving Gregoire Delétang and Rapha\"el Milli\`ere).} As Ilya Sutskever, a leading architect of today's AI models, has observed, two ideas catalyzed the recent evolution in deep learning: first, that training neural networks on the deceptively simple task of predicting the next word would drive them to \emph{compress} the data they were trained on---a thought that has since been mathematically and empirically substantiated by showing that any predictive model can be used as a compressor and any compressor can be used as a predictive or generative model \parencite[§§3.2–4]{deletang2024language}; and second, that sufficiently good compression would eventually translate into understanding \parencite{sutskever2023observation, SutskeverHuang2023FiresideChat, SutskeverMartensHinton2011GeneratingText, hutter2005_uai}. Because the patterns of human language use are, in Sutskever's phrase, a ``projection of the world onto text,'' becoming truly competent at next-word prediction requires neural networks to extract the rules governing the reality described by text: ``predicting the next token well means you understand the underlying reality that led to the creation of that token'' \parencite{PatelSutskever2023BuildingAGI}; ``If you compress the data really well, you must extract all the hidden secrets which exist in it'' \parencite{sutskever2023observation}. Sutskever illustrates the point with a detective story: accurately predicting the resolution requires more than superficial correlations---guessing ``the butler'' will not do. Accuracy demands some grasp of the particular tangle of facts, lies, and motivations involved. Part of what unlocked AI systems' striking leap in capability was thus the thesis that \emph{compression breeds broad competence}. This has brought an increasing number of computer scientists to the view that ``\emph{to understand is to compress}'' \parencite[2]{zhang2025theory}.

Yet not every compression is like Kepler's. An ingenious stenographer could rewrite Brahe's catalog in shorthand, abbreviating recurring terms to fit the entire dataset into a much shorter book. That too would be compression. But it would not be an advance in our understanding of the heavens; it would only constitute a more economical representation of the understanding that Brahe already possessed.

This contrast raises a puzzle: why does Kepler's compression illuminate the motion of the planets while the stenographer's compression seems merely to abbreviate our record of them? More generally, why are some compressions more illuminating than others? 

This puzzle is sharpened by a striking asymmetry between the different literatures on understanding. In the formal-computational tradition just surveyed, compression is treated as so central to understanding that the two are frequently identified. In the philosophical literature, by contrast, compression has played at most a marginal role. This asymmetry is significant, because philosophers of understanding have been concerned precisely with articulating what understanding consists in; if compression were simply identical with understanding, one would expect it to figure more prominently in their accounts. Their relative silence is therefore not just a gap in the literature, but a datum: it suggests either that the compressionist tradition has overidentified distinct notions, or that philosophers have been describing features of understanding whose connection to compression has not yet been made explicit.

The philosophical literature has tended to begin from the notions of \emph{knowledge} and \emph{explanation}, situating understanding in relation to them \parencite{friedman1974explanation, kitcher1981unification,Zagzebski1996-ZAGVOT-3, elgin1996judgment}. The focus has accordingly been on characterizing the distinctive epistemological and behavioral profile of understanding and the abilities it confers. Understanding has been described as \emph{grasping connections} between known facts \parencite{Wittgenstein1953-WITPI-4, kvanvig2018knowledge, grimm2011understanding, riggs2003balancing, belkoniene2023}, especially \emph{explanatory} connections \parencite{khalifa2017scientific, Strevens2013-STRNUW}, and as conferring an ability to handle novel and counterfactual cases \parencite{grimm2011understanding, hills2015understanding}, make reliable predictions \parencite{de_regt2017scientific}, and give principled explanations \parencite{hills2015understanding, deRegt2009-DERTEV}.\footnote{For overviews, see \textcite{baumberger2017understanding} and \textcite{grimm2021stanford}.} Why should compression seem indispensable from the computational point of view, yet peripheral from the philosophical one?

The unificationist tradition in the philosophy of science might be regarded as an exception, since it comes close to the compression thought by highlighting the connection between understanding and unifying explanations—see \textcite{friedman1974explanation}, \textcite{kitcher1981unification}, and especially \textcite{Schurz1994-SCHOOA}, who argue that understanding involves reducing the number of independent assumptions needed to explain a body of phenomena. But it is only with \textcite{wilkenfeld2018compression} that the information-theoretic notion of compression explicitly enters the philosophy of understanding. Wilkenfeld characterizes understanding \emph{as compression}. He argues that a person understands a phenomenon to the extent that she possesses a representation/process pair generating useful information about it at a smaller description length. But he leaves the formal-computational tradition outside philosophy unaddressed, and the question of how the two pictures might be integrated within a single account of understanding remains open.

This paper develops such an integrative account through three interlocking theses. The first thesis is what we call the \emph{Proxy Hypothesis}: the concept of understanding functions as an efficient proxy for robust competence. Human social life depends on a continual division of epistemic and practical labor that requires us to identify trustworthy partners and effective teachers, often well before we have any chance to witness their competence. The concept of understanding meets this need by tracking a particular ground of robust competence that can be probed for more efficiently than the competence it enables, making it uniquely suited to the social tasks of identifying \emph{whom to trust} and \emph{whom to learn from}.

The second thesis specifies what cognitive state the concept tracks. To understand a domain, we argue, is to possess a \emph{mental model} of its relational structure---an internal representation that captures the connections, dependencies, and regularities binding the domain together. Such a model enables prediction: it makes the domain less surprising to its possessor. And what enables prediction enables compression, because what becomes predictable no longer needs to be stored separately. But understanding breeds compression without being identical with it. Compression is the representational shadow cast by a mental model that maps relational structure. This vindicates the claim that there is a close connection between comprehension and compression while explaining why it fails as an identity claim, thereby making room for the cluster of constitutive characteristics that philosophers have foregrounded. This allows us to understand the formal-computational literature and the philosophical literature as circling the same phenomenon from different sides.

But compression comes in degrees; and the third thesis is that part of what drives the particularly high degree of compression that human understanding aspires to is the \emph{sociality} of understanding. The boundedness of individual human cognition already favors forms of understanding that fit within finite memory and remain manipulable in thought. But it does not yet explain why human understanding aspires to small sets of statable principles or compact systems that a teacher can pass on to a student in one sitting. This drive toward principled simplicity, we argue, comes from the social functions of the concept of understanding identified by the first thesis. To earn trust, understanding must be \emph{demonstrable}, i.e. take a form that can be quickly displayed in conversation. To be passed on to others, it must be \emph{transmissible}, i.e. capable of passing through the narrow bottleneck of communication.\footnote{Following recent work on the sociality of belief \parencite{ChrismanForthcoming-CHRTSA-6}, one could even conceptualize the sociality of understanding more radically, as the idea that shared understanding is prior to individual understanding, and that even private individual understanding is best understood as a potential contribution to shared understanding. But our argument here does not depend on this strong reading of the sociality of understanding.}  Understanding gets compressed for social reasons.

A distinctive feature of our methodology is that we let our model of \emph{what understanding is} grow out of our model of \emph{why we care about understanding}.\footnote{We draw inspiration from the increasingly popular methodology of reverse-engineering the most basic functions of our concepts and practices \parencite{craig1990, williams2002, price2011, price2013, fricker2016, WilkenfeldPlunkettLombrozo2018, Hannon2019-HANWTP-5, kuschmckenna2020, queloz2021, lawlor2023, thomasson2025, kelley2025}.} But there are two sides to this. One is what drove us to \emph{attribute} understanding; the other is what drove us to \emph{acquire} understanding. The attribution question asks after the development and function of the \emph{concept} of understanding. What practical needs led us to form and retain a concept which renders us sensitive to the distinction between “understanders” and “non-understanders”? The acquisition question asks after the development and function of the \emph{state} of understanding. What practical pressures encourage, in each generation anew, the formation of the kind of cognitive organization that the concept of understanding picks out and imbues with significance?

Where such questions have been addressed at all, it has tended to be in a one-sided way, focusing either exclusively on the concept or on the state of understanding.\footnote{See, e.g., \textcite{Woodward2003-WOOMTH}, \textcite{Grimm2012}, \textcite{Hannon2019-HANWTP-5}, and \textcite{Nado2025}.} A guiding assumption of our inquiry, by contrast, is that these two sides of the issue must be addressed together, because the relationship between the \emph{concept} of understanding and the cognitive \emph{state} it picks out is not one-sidedly representational, but a \emph{co-evolutionary feedback loop}. The concept could only gain a foothold by tracking a cognitive state that our capacities and circumstances made attainable and useful: explaining the concept therefore requires explaining the state. Conversely, by giving a name and value to the state we call ``understanding,'' we create an ideal that shapes our cognitive aspirations and our pedagogy---a version of what \textcite{Hacking1995-HACTLE-2} calls a ``looping effect,'' where classifications of human kinds feed back to alter the people classified. This loop is the channel through which sociality leaves its imprint on the form of understanding we gravitate toward: by giving the concept of understanding a central role in the social practices of trust and transmission, we make the cognitive state it tracks answerable to those practices, driving it toward principled simplicity.

Our project extends the unificationist and Wilkenfeldian precedents in three respects. First, we connect the philosophical question of understanding to a much wider tradition ranging across information theory, machine learning, and recent AI research---where the strongest versions of the identity claim are to be found, and where the question's contemporary stakes are most salient. Second, we resist the identity claim. Where Wilkenfeld defends understanding \emph{as} compression, we argue that the link between compression and comprehension is tight without amounting to identity---a difference we capture by treating compression as the shadow of comprehension. Third, and most importantly, we foreground the sociality of understanding and link it to compression. Where the unificationists and Wilkenfeld both note in passing that compressed understanding has communicative or social benefits, we argue that the form characteristically taken by human understanding is itself \emph{shaped by} the demands of demonstrating competence and transmitting it. This also helps us to make sense of the inscrutability of current AI systems: a system trained to optimize for predictive accuracy alone, unconstrained by the bottlenecks of demonstration and transmission, will gravitate toward what we shall call \emph{opaque} understanding: high predictive power coupled with illegible complexity. The kind of understanding humans prize—elegant, principled, and communicable---is the kind that typically emerges when compression is shaped by sociality.

We proceed as follows: \S2 develops the hypothesis that the concept of understanding serves as a proxy for a distinctive form of robust competence, helping us identify whom to trust and whom to learn from. \S3 traces the practical pressures this exerts on the cognitive \emph{state} of understanding and articulates what we call the \emph{Competence through Predictive Compression} (CPC) framework. \S4 takes up objections and \S5 draws out the lessons for machine understanding.

\section{The Functions of the Concept of Understanding}\label{the-functions-of-the-concept-of-understanding}

\subsection{Identifying Whom to Trust and Learn From}\label{identifying-whom-to-trust-and-learn-from}

What does the concept of understanding help us achieve? We suggest that its primary function is to track \emph{robust competence}: the stable, flexible, and reliable ability to succeed in performing certain tasks across a variety of situations. As members of societies that depend on a division of labor, we frequently rely on others to do
things on our behalf. This means that we constantly need to decide whom to trust. Can I rely on the pilot to fly the plane? Should I trust the doctor’s recommendation? Locating and assessing competence in performing certain tasks is an ineluctable concern in collaborative societies. And one function of the concept of understanding is to help us identify people we can trust. When a person displays the marks of understanding a domain, we take this as evidence that they can be relied upon within it. This aligns with Linda Zagzebski's observation that ``when we want an expert about a problem, we consult a person who has \emph{understanding} of the subject matter, since such a person is likely to be a reliable problem solver'' (\citeyear{zagzebski2001recovering}, 245; see also \cite{Nado2025}).

The concept of understanding therefore performs a \emph{fiduciary} function (from Latin \emph{fiducia}, trust). It helps us distinguish competence that is likely to survive novelty from competence that collapses outside familiar conditions. Someone who performs well in routine conditions is already useful, but someone who understands can be expected to handle unfamiliar and difficult conditions as well. This is why understanding is closely tied to warranted deference \parencite{wilkenfeld2016depth}.

The concept also marks out agents from whom robust competence can be acquired. This connects with Michael Hannon's (\citeyear{Hannon2019-HANWTP-5}) thesis that the point of the concept of understanding is to flag good explainers. When people display the marks of understanding
 something, we take this as evidence that they make good teachers on the
 subject. As legions of long-suffering students can attest, that remains
 a defeasible inference: some of the least comprehensible lectures have
 been delivered by the most comprehending minds. Nevertheless, offering
 principled explanations is not just a way to signal that one can be
 trusted. It is, above all, a means of \emph{transmitting} understanding
 from one agent to another; when things go well, the principles
 \emph{exhibited} in the explanation are also \emph{transferred} from the
 explainer to the addressee of the explanation.

 The concept of understanding thus also performs a \emph{transmission}
 function by enabling us to track robust competence. Explanation instills
 understanding, thereby contributing to the spread of robust competence.
 And the concept of understanding supports this transmission by helping
 us decide \emph{whom to learn from}. In this respect, understanding is socially generative: it does not merely support successful action by its possessor, but enables competence to spread.

The fiduciary and transmission functions are two applications of a single social need: the need to efficiently locate robust competence. In the fiduciary case, we locate robust competence in order to rely on it. In the transmission case, we locate robust competence in order to acquire it. In both cases, however, the immediate aim is to identify people who possess it. The concept of understanding, we hypothesize, arose as an efficient proxy for tracking robust competence.

This tight link between understanding and robust competence is also visible from the negative case. When an agent succeeds only under routine conditions and fails erratically once these vary, we take this brittleness as evidence against understanding. Recent debates about machine understanding have brought this out: Mitchell and Krakauer write that ``the oft-noted \emph{brittleness} of these AI systems---their unpredictable errors and lack of robust generalization abilities---are key indicators of their lack of understanding'' (\citeyear{mitchellkrakauer2023}, 1). This inference from brittleness to lack of understanding makes sense if the concept of understanding emerged to track robust competence.

\subsection{The Proxy Hypothesis}\label{the-proxy-hypothesis}

 Given this tight link, one might be tempted to simply \emph{assimilate}
 understanding to robust competence. Robert Brandom, for instance, argues
 that understanding just \emph{is} practical mastery. A student seeking
to understand mathematics must not wait for some ``inner light'' to go
 on, but simply ``practice making the moves \ldots{} until he masters the
 practical inferential abilities in question'' (\citeyear{Brandom2009-BRARIP}, 172--3).
 Similarly, Philipp Koralus contends that ``what is worth caring about in
 the notion of understanding'' is ``\emph{competence}'' (\citeyear{koralus2022}, 35). He
 concludes that it is competence we should focus on going forward, both
 in the context of human reason and in the context of artificial
 intelligence.

Yet assimilating understanding to competence collapses a distinction we have reason to preserve. Competence is a \emph{behavioral} property, manifested in what an agent succeeds in doing across the tasks she undertakes. Understanding is a \emph{cognitive} property, pertaining to how her mind is organized. The distinction matters because robust competence is difficult to track directly. We usually observe only a thin slice of past performance, and often only under routine conditions. Assembling that track record requires a significant investigative effort even while its value for predicting future success is limited, since what looks like stable competence may prove fragile once enough variation is introduced.

This creates an epistemic challenge: we have a pressing practical need to identify a behavioral
property—robust competence—that is (a) labor-intensive to keep track of, (b) only partially observable,
and (c) easily confused with something more brittle. The practical task of identifying whom to trust and whom to learn from therefore requires a more tractable and reliable signal than past performance alone can provide.

We hypothesize that the concept of understanding developed as our solution to this challenge. The
concept serves as a reliable and efficient conceptual proxy for tracking robust competence. Call this the
\emph{Proxy Hypothesis}. Given the infeasibility of an exhaustive and definitive assessment of robust competence,
we largely track it indirectly, by tracking a cognitive property we call ``understanding.''

What cognitive property is this?  ``Comprehension,'' the Latinate twin of the Germanic term ``understanding,'' offers a clue: deriving from \emph{com-} (``together'') and \emph{prehendere} (``to seize''), it evokes a ``grasping together,'' as if unifying various elements in a single clasp. Several philosophical accounts converge on this picture. Ludwig Wittgenstein speaks of understanding in terms of ``seeing connections,'' (\citeyear{Wittgenstein1953-WITPI-4}, §122). Jonathan Kvanvig argues that ``understanding seems to involve conceptual and explanatory connections between various items of information that are seen or grasped by the person in question'' (\citeyear{kvanvig2018knowledge}, 697). To grasp such connections is to see ``how things hang together'' \parencite{riggs2003balancing} or ``how various parts relate to each other'' \parencite{grimm2011understanding}. What the concept of understanding flags, then, is a \emph{grasp of relational structure}—of the connections between parts of an object, a situation, or a subject matter. 

\begin{figure}[h]
    \centering
    \begin{minipage}{0.54\linewidth}
        \centering
        \includegraphics[width=\linewidth]{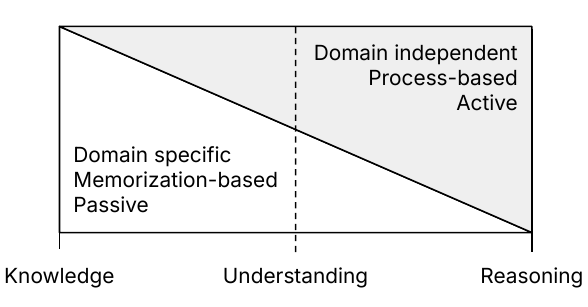}
    \end{minipage}%
    \hspace{0.35cm}
    \begin{minipage}{0.38\linewidth}
        \caption{The concept of understanding marks the crossover between knowledge and reasoning.}
        \label{fig:und_crossover}
    \end{minipage}
\end{figure}

Understanding thus reaches beyond the possession of disconnected facts. It is not exhausted by knowledge, which can consist of a passive repository of isolated true beliefs, nor by reasoning, which is active, but can remain abstract and domain-general. As Figure \ref{fig:und_crossover} illustrates, understanding occupies a middle ground between knowledge and reasoning: a person understands a domain when she integrates domain-specific knowledge into a relational structure that allows her to handle novel cases and thus actively generate new insights.

A novice driver might know a few isolated rules, such as:
``If the temperature gauge is in the red, pull over.'' But a skilled
mechanic truly understands the engine because they have grasped the
underlying connections: they know the radiator cools the engine fluid,
which circulates to prevent the engine from overheating; they know a
faulty thermostat can block this circulation, causing the temperature to
rise. The mechanic has grasped the causal and functional relationships
between the parts, which in turn empowers them to predict what
intervention will fix the problem. Similarly, when evaluating a doctor's
competence, we are less interested in their ability to recall isolated
trivia about diseases and more interested in their grasp of essential
connections and dependencies between causes, symptoms, and treatments.
Those who grasp how things are connected can apply their understanding
to new cases and counterfactuals and make better predictions about them
\parencite{hills2015understanding, de_regt2017scientific}.

The connections most crucial for securing this robust competence are
\emph{explanatory} ones \parencite{khalifa2017scientific, baumberger2017understanding}. These can be,
for instance, \emph{causal-mechanical} connections that explain the
physical processes of a phenomenon \parencite{grimm2006species, pritchard2014} or
\emph{law-making} connections that unify disparate events under a
general principle \parencite{friedman1974explanation, kitcher1981unification, belkoniene2023}. But grasping other types of
connections, such as logical, probabilistic, or conceptual ones, can also be 
sufficient for certain forms of competence \parencite{kvanvig2003value, kvanvig2018knowledge}.

But what exactly does ``grasping'' such connections consist in? We propose that it consists in forming a \emph{mental model} of the domain: what Kenneth Craik \parencite*{Craik1943} called a ``thought model,'' namely an internal representation that exhibits a ``similar relation-structure'' \parencite[51]{Craik1943} to the process it models.\footnote{See \textcite{koralus2022} for an illuminating discussion of Craik's thought models.} Such models may be explicit schemas, diagrams, causal models, or entire principled systems; but they can also, up to a point, remain tacit, as settled patterns in one's perceptual sensitivity and inferential dispositions (we return to this contrast below). What matters is that they preserve enough of the represented domain's relational structure to support prediction, explanation, and counterfactual reasoning.

If we think of understanding as the state of possessing a structure-sensitive mental model of a domain, its connection to robust competence becomes visible. Because such a model preserves relational structure across variation, it enables its possessor to extrapolate beyond familiar situations and anticipate what would happen in counterfactual circumstances. The formation of a structure-sensitive model thereby grounds a distinctive form of robust competence: competence that travels beyond the cases through which it was acquired.

Not all forms of robust competence are grounded in mental models. Achilles may be robustly competent in battle, but much of what grounds his success lies in his reflexes, muscle memory, and bodily characteristics such as strength, speed, and endurance; a mental model of the dynamics of combat may play some role in his exploits, but it is not the core of what makes him formidable on the battlefield. Kepler offers a very different case: his robust competence in astronomical prediction rests almost entirely on his grasp of the elliptical structure of planetary motion. Physical prowess, muscle memory, cultivated character, and structure-sensitive cognitive organization thus all figure among the possible grounds of robust competence. Robust competence is \emph{polygenic}: it can be grounded in different underlying properties, and these may be mixed to varying degrees.

This plurality of grounds is mirrored in the plurality of our concepts for tracking robust competence. We speak of an agent's \emph{strength}, \emph{skill}, \emph{virtue}, and \emph{understanding} to register distinct forms of competence-grounding, each with its own conditions of ascription and its own practical relevance. Such a division of conceptual labor allows us to attend to the particular ground that matters most in a given practical context. When we want to identify someone who can help us move heavy furniture, we attend to strength. When we want to determine who will make a reliable diagnostician or an illuminating teacher, we attend to understanding. Each concept refines our sensitivity to one route by which robust competence can come about. Understanding is the concept that tracks one such route: the possession of a structure-sensitive mental model.

Evaluating the robust competence of agents in terms of whether they
possess understanding is simultaneously less labor-intensive and more
reliable than relying on track record alone. It is less labor-intensive, because even without a track
record of past performance, we can rapidly assess how robustly competent
someone is by probing for understanding: we can ask them to summarize or
explain the principle behind something, probe their ability to handle a
counterfactual, or give them a particularly challenging test case.

This is also more reliable, because two
agents can have the same track record, yet differ in their cognitive
organization in ways that matter for counterfactual and novel cases:
only an agent with a real grasp of relational structure can be relied
upon to continue to draw the right inferences. We know that someone
might obtain perfect scores on an exam and still lack understanding.
Narrow competence on a test set does not entail \emph{robust}
competence. Indeed, one of the challenges of exam design is to put
together a set of tasks that test for genuine understanding \emph{as
opposed to} the fragile competence achievable through rote memorization.
An assessment of understanding is therefore never just about the task at
hand; it is always also an assessment of whether the agent's competence
will \emph{transfer} to novel and counterfactual tasks. When we judge
that a student has not only given the correct answer, but truly
understood the principle underlying it, we are implicitly predicting
that they would be able to solve a host of different problems as well.

This forward-looking aspect, together with greater efficiency, is the
key advantage that appraisals of understanding have over mere appraisals
of past performance. Hence the merit of tracking robust competence
\emph{via} the concept of understanding (\emph{in addition} to keeping track of
past performance---the two are not exclusive).

By serving as an efficient and reliable proxy for the model-based ground of robust competence, the concept of understanding thus fulfills both its fiduciary function of helping us identify whom to trust, and its transmission function of helping us identify whom to learn from.

However, as our co-evolutionary feedback loop hypothesis suggests, these
social functions do more than justify the concept's existence. They form
the environment that shapes the cognitive state of understanding itself.
We now turn to the other side of this loop: the practical pressures that
these social demands place on the formation and character of the
cognitive state of understanding.

\section{How Sociality Compresses Comprehension}\label{how-sociality-compresses-comprehension}

\subsection{Four Pressures Shaping
Understanding}\label{four-pressures-shaping-understanding}

We now use the social-epistemic functions of the \emph{concept} of understanding as a guide to the pressures shaping the \emph{state} of understanding. If the concept helps us identify agents whose competence can be trusted and transmitted, then the cognitive state it tracks will be shaped by the conditions under which competence can be reliable, recognizable, and shareable.

The first pressure is \emph{predictivity}. Robustly competent agents must be able to anticipate what is likely to happen: successful action, diagnosis, explanation, and intervention all depend on expectations about how a situation will unfold. A structure-sensitive model grounds robust competence because it makes the domain less surprising to its possessor. It enables the agent to anticipate what will follow, what would happen under variation, and which interventions are likely to matter.

On its own, predictivity might favor ever more complex models, provided that added complexity improved accuracy. Human cognition, however, is finite. Models must remain \emph{storable} within the bounds of memory and \emph{manipulable} in thought. A model too complex to retain or use would not support robust competence in creatures like us. The second pressure is therefore toward \emph{storability}-\emph{cum-manipulability}.

These two pressures already point to the need for mental models that
are powerfully predictive, yet economical. But storability and
manipulability require that a model not be overwhelmingly complex; they
do not require the model to be particularly elegant or concise. To
understand where the even stronger pressures toward simplicity that
give human understanding its distinctive character come from, we must
factor in the social dynamics of understanding. This reveals two further
pressures.

The third pressure is \emph{demonstrability}. Since robust competence is only indirectly observable, agents need reliable and efficient signals of whom to trust. This creates a social pressure on understanding to become displayable. If understanding is to be demonstrated quickly and effectively, it cannot remain an inscrutable thicket of ineffable sensitivities. It must be capable, at least in favorable cases, of being distilled into statable, consistent, and coherent principles. Such principled organization enables concise explanation, and the fewer and more general the principles, the more elegant the understanding one can display.

The social need for understanding to be demonstrable thus drives understanding toward principled simplicity. The expert who wants to advertise her competence needs her understanding to be concisely displayable in the form of a few explicit principles. More broadly, those whom others rely on will often be expected to vindicate the trust they receive by demonstrating their
understanding.\footnote{This aligns with Belkoniene's emphasis on the fact that understanding must be based on ``reflectively accessible grounds'' \parencite[349]{belkoniene2022}.} A predictive but incommunicably complex form of
understanding would fail to meet this crucial social need, as it
cannot easily be exhibited to earn and justify trust.

The fourth pressure is \emph{transmissibility}. Because one central reason to track understanding is to identify whom to learn from, understanding must be capable of passing through the narrow bottleneck of communication. It is one thing to display just enough of one's mental model to earn trust, as a physician might do when quickly explaining a diagnosis to a patient. It is quite another to transmit the entire model so as to reproduce the competence it confers, as a teacher aims to do with students. The most effective teachers are often those who have distilled a domain down to its basic principles and arranged them in a coherent, memorable system. Thus, the social pressure toward transmissibility is another driver toward principled simplicity. As creatures who do not merely acquire understanding individually, but share it with others, we are driven toward mental models that are elegantly principled and communicable.

\begin{figure}[h]
    \centering
    \includegraphics[width=0.7\linewidth]{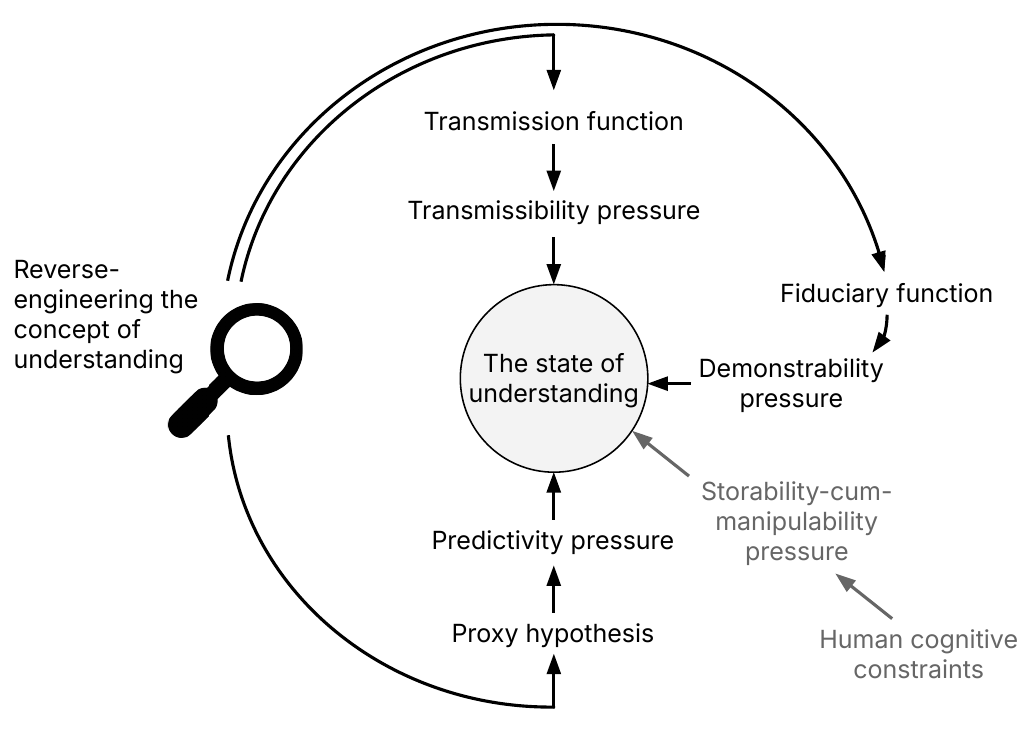}
    \caption{The four pressures shaping the state of understanding.}
    \label{fig:pressures}
\end{figure}

As summarized in Figure \ref{fig:pressures}, this gives us four distinct pressures molding
the state of understanding. The social-epistemic function of the concept therefore reveals pressures that would remain invisible to an individualistic theory of cognition. The need to identify trustworthy experts and effective teachers favors a particular \emph{kind} of understanding: understanding that is not merely predictive, but also efficiently communicable. The more elegantly principled understanding is, the more efficiently it can be demonstrated and conveyed. This is the co-evolutionary feedback loop in action: the need to acquire understanding favors a concept for recognizing it in others, while the social use of that concept refines the shape of the state itself. Sociality is a key driver of principled simplicity.

\subsection{From Comprehension to Compression}\label{from-comprehension-to-compression}

How can a mental model simultaneously satisfy the demands for predictivity, storability, demonstrability, and transmissibility? We argue that these pressures converge in favoring \emph{predictive compression}: the formation of models that reduce surprise by capturing relational structure while remaining economical in representation. Our \emph{Competence through Predictive Compression} (CPC) framework treats this as the functional core of how understanding grounds robust competence: it is a cognitive organization that confers robust competence \emph{through} predictive compression.

The path from comprehension to compression runs through prediction. The very features that render a model predictive also enable it to compress the data it describes.\footnote{A point echoed in recent work in machine intelligence research---see \textcite[7]{ramstead2025noumenallabswhitepaper}.} As Jürgen Schmidhuber puts it, ``whatever you can predict you can compress as you don't have to store it extra'' (\citeyear{schmidhuber2010}). If one can predict something well, this means one has discerned a \emph{pattern} in it. It is not random; it exhibits some regularity. That pattern can then be used to describe the data more economically, because ``a pattern in some data is any regularity that allows the lossless compression of those data'' \parencite[77–78]{Petersen2026}. Given a random string of letters, for example, one cannot predict what comes next and must store the whole string. But if the string follows a pattern, such as ``ABABAB,'' one can predict the next letter at any point. This allows one to compress the entire string into the rule: ``Repeat AB.''

The link from better predictions to shorter descriptions is forged by a
fundamental insight from information theory: the information content of
something can be quantified in terms of how \emph{surprising} it is. As
Claude Shannon established, an event that is highly probable is not
surprising, and being told it occurred provides little information
(``The sun rose this morning'' is not news). An improbable event, by
contrast, is highly surprising and conveys a great deal of information.
The goal of a predictive model can be framed as an effort to minimize
its average surprise at incoming data. A good model is one that assigns
high probabilities to events that actually transpire and low
probabilities to those that do not.

This provides the bridge to compression. A string of data is truly
\emph{random} if and only if the information required to describe (or
transmit) the series is \emph{incompressible}: nothing shorter than a
verbatim rendition of the entire string will preserve it. Conversely,
the string has a \emph{pattern} if and only if it admits of a
description that is shorter than the string itself \parencite{chaitin1975, dennett1991, ladyman2007, sune2021}. To discern redundancies, repetitions, or even just uniformly colored
areas is already to discern patterns, enabling shorter descriptions by
abbreviating the predictably repeated strings or encoding uniform areas
by specifying only their boundaries and color (as in
``color-by-number'' books) instead of describing them pixel by pixel.
Similarly, once one understands that certain events are more likely than
others, one can use shorter descriptions or shorthand symbols for them.
Even the realization that certain letters are more frequent than others
enables compression; one gives common letters short encodings and rare
letters longer ones (this is done in Huffman coding, for example). The
better a model forecasts the data, the shorter the description needed to
specify that data \emph{given} the model. The payoff of good prediction
is compression.

The same point holds for understanding the world more widely. Imagine you must send a message each hour saying whether it is raining (``R'') or dry (``D''). A naive baseline treats each hour as independent and requires 24 symbols per day. Now suppose you discover a pattern: when a cold front arrives, it brings a six-hour band of rain. Because you expect contiguous rainy hours once the front appears, you can encode the day's sequence with fewer symbols by giving the starting hour of the band plus any deviation from the rule. Prediction enables compression because structure makes data less surprising.

More generally, encoding stable relations into a model lowers average surprise across new data. A law, symmetry constraint, conservation constraint, or causal dependency prunes away large regions of the space of possible outcomes, thereby making certain observations systematically less surprising. This is compression \emph{through sensitivity to structure}. The representation is shorter than the data because it exploits the relational structure of the domain. This is why it is common for theorists outside philosophy to treat compression as the hallmark of understanding: compression is the representational shadow cast by structure-sensitive prediction.

However, structure-sensitivity itself comes in degrees reflecting the
\emph{depth} of one's understanding, because one can be sensitive to
structures of varying depth. An illustration of this spectrum is the
progression in understanding planetary motion realized by Tycho Brahe,
Johannes Kepler, and Isaac Newton.\footnote{On Brahe, see \textcite{thoren1991}. On Kepler, see \textcite{voelkel}. On Newton, see \textcite{westfall2015}. On the
  progression in understanding they made possible, see \textcite{li2021kepler}.}

Brahe recorded the motions of the planets with unprecedented precision,
thereby acquiring a grasp of their ``relational structure'' in the most
superficial sense: he provided snapshots of how they spatially related
to each other at different moments. But he remained unable to
\emph{predict} the movements of the planets, because he had not grasped
the underlying patterns---the relational structure in the deeper sense
of hidden regularities.

Kepler discerned these underlying patterns and articulated them in his
three laws of planetary motion: planets move in elliptical orbits with
the Sun at one focus, sweep out equal areas in equal times, and the further a planet is from the Sun, the more slowly it completes its orbit. This enabled him to predict planetary motion within a
small fraction of a degree and facilitated compression: given a few
observations about Venus's position, Kepler could work out its position
on subsequent days. But he could not explain \emph{why} the planets
moved according to the patterns he had discerned.

It was only with Newton, who formulated the laws of motion and universal
gravitation underpinning these patterns, that an even deeper
understanding of the underlying structure was provided. Newton grasped
the relational structure not just in the sense of discerning hidden
regularities, but in the sense of capturing the underlying laws. And
with Newton's laws unifying our understanding of terrestrial and
celestial mechanics, the movements of moons and comets became
unsurprising as well.

The progression from Brahe through Kepler to Newton also goes hand in
hand with greater compression and easier transmission: instead of
handing over a thick book of coordinates, one need only share ``orbital
elements''---a few key parameters that characterize a planet's
orbit---such as its semi-major axis (half its longest diameter) and its
eccentricity (its deviation from a perfect circle); the laws can then be
used to generate the rest of the path.

A more formal account can be given using information theory and the minimum description length principle (see \cite{grunwald2007}). Let $D$
be a finite dataset of observations relevant to some task family. Let
$L( \cdot )$ be description length in bits. To encode $D$ using a
model, one chooses a model $M$ and then encodes the data relative to
it. The resulting description length is:
\begin{align*}
L(M)  + L(D|M)
\end{align*}

\noindent 
Think of $L(M)$  as the cost of storing the model (e.g. rules,
principles, constraints, and dependencies), and $L(D|M)$ as the cost
of storing the data \emph{given} the model, i.e. the cost of storing the
\emph{residuals}, such as initial position, deviation from the rule,
noise, etc. The description length of the data given the model,
$L(D|M)$, is a formal measure of the model's total surprise at
that data. Encoding stable connections---a law, a symmetry, or a causal
dependency---into one's model has the effect of making certain new data
systematically less surprising, thereby reducing $L(D|M)$.

Compression is achieved if the total description length of the model
plus the data given the model is smaller than a baseline without a
model, e.g. storing $D$ as raw data. We say that $M$
\emph{compresses} $D$ iff
\begin{align*}
L(M)  + L(D|M) < L(D)
\end{align*}

\noindent 
But note that when ``data are compressed'' in this sense, they are not
literally crammed \emph{into} a model; rather, the description of the
data is split into two parts: the model $M$ which is the part that
encodes reusable \emph{structure}; and the data as encoded under $M$,
which contains only residuals, i.e. whatever is left over once one has
accounted for the regularities captured by $M$.

Predictive compression is thus not about discarding data, but about
building a model that renders the data less surprising and therefore
easier to describe. This measure of a model's average surprise has a
formal name: \emph{negative log-likelihood} (or \emph{cross-entropy}).
The core idea is to quantify the surprisingness (or ``surprisal'') of a
single outcome as $-\log(P(outcome))$. This captures the inverse
relationship we intuitively expect between probability and surprise: as
an outcome's probability $P(outcome)$ approaches $1$ (certainty), its
negative logarithm approaches $0$, meaning there is zero surprise---an
event that is guaranteed to happen tells us nothing new. Conversely, as
an outcome's probability $P(outcome)$ approaches $0$ (near impossibility),
its negative logarithm grows toward infinity, reflecting immense
surprise. A cognitive system that seeks to understand a domain is
therefore engaged in a process of adjusting its internal model to
minimize this surprise on average across all the data it sees, which is
equivalent to maximizing the likelihood it assigns to that data. The
total description length of the data given the model, $L(D|M)$,
can thus be seen as the sum of all these individual surprisals. It is a
formal measure of the model's cumulative predictive error.

Putting these ideas together, we can say that mental models must solve an optimization problem: keep representations simple enough to store and share, yet structure-sensitive enough to yield robust predictions. In formal terms, a mental model is formed in pursuit of the following objective:
\[
\min_{M} \; L(D_{future} \mid M) + \lambda\, L(M)
\]

\noindent 
Call this the \emph{ultimate objective} of mental model formation. The first term, $\min L(D_{future}|M)$, aims to minimize the description length of future data under the model, and so favors models that assign high probability to what will actually occur. Call this ``data compression.'' The second term, $\min L(M)$, aims to minimize the description length of the model itself, and so favors models that are easy to store, manipulate, demonstrate, and share. Call this ``model compression.'' The parameter $\lambda$ reflects how heavily one weights compact simplicity relative to predictivity. A $\lambda > 1$ encodes a strong preference for simple models, while $\lambda < 1$ permits greater model complexity. Another interpretation of $\lambda$ would be that the parameter encodes the degree of one’s \emph{presumption of systematizability}: the expectation that a simple, unified model of the domain exists. Different contexts may call for different values of $\lambda$. A theoretical physicist searching for a unifying law may have reason to set it high; an ethnographer attending to cultural specificity may have reason to set it lower.

Because we cannot \emph{directly} optimize on as-yet-inaccessible future
data, however, agents approximate the above by learning on accessible
\emph{past} data. Accordingly, the \emph{proximate} objective (through
the pursuit of which the ultimate objective is pursued) is to minimize
the total description length of the model plus the \emph{past} data
given the model:
\begin{align*}
\min_{M} L(D_{past}|M) + \lambda L(M)
\end{align*}

\noindent 
Call this the \emph{proximate objective} of mental model formation. The risk is that an agent may rely on superficial correlations that reduce $L(D_{past}|M)$ without reducing $L(D_{future}|M)$. Hewing too closely to past data can produce overly complex models that mistake idiosyncrasy for structure. This is why predictive power must be balanced against simplicity. Increasing $\lambda$ helps steer inquiry toward structure-sensitive compression by penalizing overfitting to past data. At the same time, it favors models that are easier to store, manipulate, demonstrate, and transmit. The formal pressure against overfitting and the social pressures toward communicability converge.

\begin{figure}[h]
    \centering
    \includegraphics[width=0.6\linewidth]{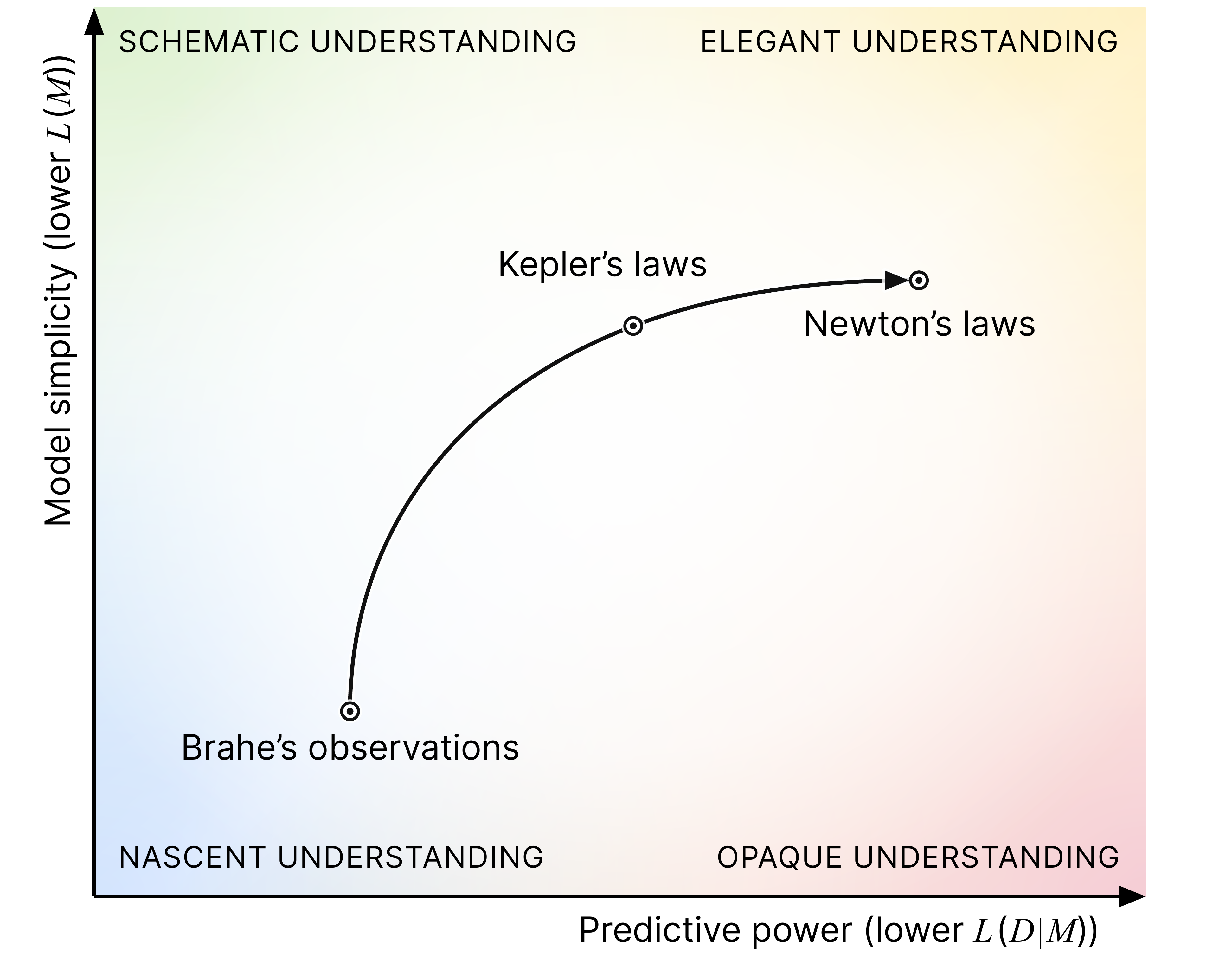}
    \caption{The space of possible forms of 
understanding mapped along two axes: model
simplicity (vertical) and predictive power (horizontal). The progression
from Brahe through Kepler to Newton illustrates a development from nascent to
elegant understanding. (The diagram is intended merely as an illustrative aid.)}
    \label{fig:space}
\end{figure}

The trade-off between model simplicity, which corresponds to minimizing
$L(M)$, and predictive power, which corresponds to minimizing
$L(D|M)$, can be visualized as a two-dimensional \emph{space of possible forms of 
understanding}, shown in Figure \ref{fig:space}. The iridescent coloring marks the four poles within that space while also indicating that these poles are connected by continuous transitions rather than by hard categorical boundaries.

On the one hand, there are strong pressures toward \emph{model
compression}---minimizing $L(M)$. These pressures, which push cognitive
systems toward the upper half of Figure \ref{fig:space}, stem directly from our
nature as bounded and social agents. 

On the other hand, there are pressures toward \emph{data
compression}---minimizing $L(D|M)$. These pressures push cognitive
models toward the right half of the map. The most fundamental pressure
here is toward predictivity. A good model minimizes our surprise at the
world by assigning a high probability to what actually happens. This
predictive success in turn allows for data compression. Because a simpler model is less likely to mistake noise
in past data for a signal, moreover, the pressure to minimize $L(M)$ 
simultaneously helps avoid overfitting, rendering models more likely to
generalize to future data and thereby reducing $L(D|M)$. In Figure
\ref{fig:space}, this means that moving up also tends to push one to the right.

The progression of scientific insight often charts a path through this
landscape. Brahe's observations represent a
form of \emph{nascent} understanding (bottom left quadrant, blue)—a complex dataset with little predictive
structure. Kepler's laws chart a path upward and to the right,
introducing a dramatically simpler model (higher on the vertical axis)
that also unlocks powerful predictive abilities (further on the
horizontal axis). Newton continues this path toward the top-right,
offering a highly compressed model that
	nevertheless yields extremely precise predictions over a vast range of
phenomena. This pairing of model
simplicity with high predictive power constitutes \emph{elegant} understanding (top-right
quadrant, yellow).

When unchecked by social pressures toward
compression, however, pushing for predictive accuracy can lead to the
bottom-right quadrant, marked in red: \emph{opaque} understanding. This is where many
contemporary deep learning systems and large-scale climate models end up
(see \cite{Knusel2020-KNSUCP, freeborn2026model}). These achieve high predictive accuracy
via vast, inscrutable models that are strong on predictivity but weak on
storability, demonstrability, and transmissibility.

And when applied in domains with low predictability---such as
history, where events are often unique and highly
contingent---we end up in the green, top-left quadrant, with \emph{schematic}
understanding. The degree of predictive compression achievable by historical understanding is modest, but
not zero, and it still consists in capturing
available structure, even if that structure is schematic rather than
nomic. A historian who understands a period may not possess a
predictive law, but they do possess compressed models in the form of
narrative templates, schemas of political and economic pressures, and
models of human motivation, which are routinely invoked by
historians to make sense of events such as the outbreak of a war or the
fall of an empire. These schemas also support counterfactual
reasoning (e.g., ``What would have happened if Archduke Ferdinand had
not been assassinated?''), indicating a grasp of causal dependencies.

Table 1 summarizes these pressures toward model and data compression
using the example of Kepler's laws:

\begin{table}[H]
\centering
\small
\begin{tabular}{|p{6.5cm}|p{9.5cm}|}
\hline
\multicolumn{2}{|l|}{\textbf{\textit{Model compression:} pressures toward minimizing $L(M)$:}} \\[2ex]
\hline
1. Pressure toward model \textit{storability} and \textit{manipulability} & Kepler's three laws are easy to remember and apply. \\
\hline
2. Pressure toward model \textit{demonstrability} & One's understanding of planetary motion can be concisely displayed by citing Kepler's laws. \\
\hline
3. Pressure toward model \textit{transmissibility} & Kepler's compact laws can be easily taught. \\
\hline
4. Pressure to avoid \textit{overfitting} & Kepler's laws extend to new planets. \\
\hline
\multicolumn{2}{|l|}{\textbf{\textit{Data compression:} pressures toward minimizing $L(D|M)$:}} \\[2ex]
\hline
1. Pressure toward \textit{predictivity} & Kepler's laws predict Venus's future positions. \\
\hline
2. Pressure toward data \textit{storability} & Venus's trajectory can be stored by storing only its orbital elements (from which the full trajectory can be retrieved). \\
\hline
3. Pressure toward data \textit{demonstrability} & One's possession of Venus's full trajectory can be concisely displayed by showing only its orbital elements. \\
\hline
4. Pressure toward data \textit{transmissibility} & Venus's full trajectory can be transmitted by conveying only its orbital elements. \\
\hline
\end{tabular}
\caption{The practical pressures driving bounded social agents
to form mental models.}
\end{table}

One feature of the framework is that a wider \textit{scope} of understanding is favored when the objective $L(D|M) + \lambda L(M)$ is applied across domains. A narrow model, such as Kepler's laws, compresses planetary positions well, but assigns high surprisal to observations of moons, comets, tides, and terrestrial motion. Newton's laws compress these domains together, achieving lower total description length when $D$ spans them all. The same objective that favors simplicity and predictivity therefore also favors wider scope: the more domains a model renders unsurprising, the lower the total length becomes.

We can now see how the account unifies two literatures that have circled the same phenomenon from different sides: the philosophical literature that treats understanding as grasping connections and the formal-computational literature that treats understanding as compression. These are not rival pictures. Grasping connections is what makes compression possible. At a functional and information-theoretic level, to grasp connections is to form a structure-sensitive model that enables compression by predicting data under that model.

This also lets us reevaluate Chaitin's dictum that ``comprehension is compression'' and explain both its attraction and its limitation. The dictum is correct if it means that genuine comprehension characteristically enables compression. It is wrong if read as identifying comprehension with compression. Compression is an expression of greater comprehension of a target domain only when that compression is achieved through sensitivity to the domain's structure---and even then, it is still worth distinguishing (i) the mental model whose possession constitutes understanding from (ii) the compression of the data that this model enables and (iii) the compression of the model itself.

The account also offers an interpretation of common computer science formalisms that reveals them to express deeply human exigencies. When we give equal weight to model simplicity and predictive fit ($\lambda = 1$), our objective function becomes an instance of the Minimum Description Length (MDL) principle, a computable application of ideas pioneered by Ray Solomonoff \parencite*{solomonoff} and Andrey Kolmogorov \parencite*{kolmogorov}. What may look like an abstract principle from information theory is revealed to be a formal expression of concrete practical pressures of human life.

Similarly, the model-complexity term $\lambda L(M)$, which the machine learning literature regards
primarily as a mathematical regularizer designed to prevent overfitting, turns out to have a richer
set of rationales in the context of human learning. Our account shows that for human beings, this
term also represents something more concrete: the practical pressures of cognitive finitude and social
codependence. We are not merely optimizing for predictivity; we are optimizing for predictivity with
models that can be stored and shared.

\subsection{\texorpdfstring{What Constitutes and What Indicates
Understanding
}{What Constitutes and What Indicates Understanding }}\label{what-constitutes-and-what-indicates-understanding}

Based on this CPC framework, we can now (a) articulate the constitutive conditions for understanding, (b)
derive the indicators we rely on to guide our attributions of understanding, and (c) clarify the relationship
between understanding and explaining.

On our account, understanding is neither a distinctive feeling nor reducible to a purely behavioral property like competence. It is a cognitive organization that explains why certain forms of competence are robust. We define it as follows:

\begin{quote}
\begin{tcolorbox}[abstractbox]
\textbf{Understanding (Def.):} A cognitive system \emph{understands} a domain iff it possesses a structure-sensitive mental model that facilitates the prediction and compression of data in that domain. The \emph{depth} of understanding can be graded accordingly: deeper understanding corresponds to a model that is more compressed, more accurate (i.e. it assigns lower surprisal to new data), and applicable across a wider range of data.
\end{tcolorbox}
\end{quote}
This is what it is to understand, on our account. But the constitutive condition is not directly observable. In practice, we rely on a cluster of \emph{indicator properties} to tell whether understanding should be attributed. The CPC framework suggests that the most important indicators will include:

\begin{itemize}
\item
  \textbf{Principled Explanation}: The ability to concisely explain
 by explicitly articulating underlying principles externalizes a
  compressed model (low $L(M)$). A good explanation strips away incidental
  details and highlights the core relational structure.
\item
  \textbf{Accurate Prediction}: The ability to forecast novel data is
  the most direct evidence of a mental model that minimizes
  $L(D_{future}|M)$.
\item
  \textbf{Success on Hard Cases}: The ability to deal with particularly
  challenging, previously unseen problems. This tests whether the model
  is truly structure-sensitive and not overfitted to past data.
 \item
   \textbf{Handling Counterfactuals}: The ability to answer ``what-if'' questions and predict, in a principled and coherent way, how something would behave if
conditions were different.
\item
  \textbf{Patterned Error}: Understanding does not imply the absence of
  error, but that even one's errors exhibit a pattern, systematically reflecting the limits of the model's core assumptions or
  idealizations. By contrast, an agent without understanding fails
  erratically.
\end{itemize}

The importance of principled and concise explanation as an indicator of understanding is a direct
reflection of its co-evolution with the concept of understanding; we value the ability to explain concisely
because our concept was forged in large part to identify those who could effectively signal and transmit
their competence. This accommodates, but strictly speaking \emph{inverts} Michael Hannon's (\citeyear{Hannon2019-HANWTP-5}) hypothesis that the point of the concept of understanding is to identify good explainers. Our hypothesis is rather that the concept
of understanding fundamentally serves to track robust competence by flagging agents that have achieved
predictive compression through mental models, and \emph{one way to do this} is to look for good explainers. In
humans at least, the ability to explain things well indicates that someone has achieved structure-sensitive
compression, which is in turn a good predictor of robust competence. Moreover, once we have identified
agents with understanding, the understanding they possess can be transmitted through explanation. This social multiplication of the value of understanding further encourages the formation of mental models that can more easily be transmitted.

These indicators are fallible, but they are not arbitrary. They work because it is difficult, at least for human agents, to consistently explain, predict, handle counterfactuals, succeed on hard cases, and err in patterned ways without possessing the mental models that constitute understanding.

With this characterization in place, we can also sharply distinguish understanding from explaining---a distinction that risks being blurred in literature focused on the analytic structure of good explanations (e.g. \cite{Strevens2013-STRNUW, Khalifa2012-KHAIUO, deRegt2009-DERTEV}). De Regt (\citeyear{deRegt2009-DERTEV}, 25), for example, claims that to understand a phenomenon \emph{just is} to have an explanation of that phenomenon. On our framework, understanding is the state of possessing a mental model, while explanation is, in the first instance, a social and communicative act: the speech act aiming to transmit a mental model to another agent.

While other senses of ``explanation'' exist (a proposition can explain another without any communicative act), the sense most directly illuminated by our account is thus the one in which explaining is an attempt to transfer a mental model.\footnote{This aligns with Turri's \parencite*{turri} thesis that the norm of (the speech act of) explanation is to \emph{express} understanding and with the claim that the practice of explanation is unified by the function of disseminating understanding \parencite{Gaszczyk2025}.} In information-theoretic terms, an explanation of a target event $T$ is a communicative attempt to instill in another agent a model $M$ capable of generating the data $D_T$ relevant to $T$. It succeeds when the receiver forms such a model and can use it to reduce the surprisal of $D_T$. Explanations are the principal
social technology for moving these models between minds.

\section{Objections and Replies}\label{objections-and-replies}

\subsection{Compression Without Comprehension?}\label{compression-without-comprehension}

In response to the CPC framework's claim that compression is enabled by comprehension, readers may object that compression can occur without comprehension. A program that creates a \texttt{.zip} file, for example, achieves compression, but surely one would not say that it understands the text it compresses. Does this not show that there can be compression without comprehension after all?

\textit{Reply:} It is true that creating a \texttt{.zip} file does not by itself amount to understanding \emph{what the compressed text is about}. But ordinary file compression is not wholly detached from structure-sensitivity. A \texttt{.zip}-style compressor shrinks a file by finding repetitions and by noticing that some symbols occur more often than others. In this modest sense, the compressor has expectations about what is likely to come next. This is why Delétang et al. can turn a standard lossless file compressor into a crude predictive model and use it to generate data \parencite[\S3.4]{deletang2024language}. The generated output is noisy, because the expectations are shallow---the compressor has not identified the subject matter of a text or the objects in an image; but it has latched onto real patterns in the recurrence of strings and the frequency of symbols.

This also lets us take a more nuanced view of the stenographer case from the introduction. The stenographer who compresses Brahe's catalog in shorthand need not understand anything about the planets. But neither is this compression wholly unrelated to comprehension. To compress the catalog successfully, the stenographer must discern and exploit patterns in Brahe's records. The difference between Brahe's and the stenographer's compression lies rather in what predictive power they each get from it. The stenographer's technique may generalize across texts that share a notation, but it will not support predictions about the position of Venus. The case therefore does not show that compression can occur without understanding. What it shows is that compression is \emph{promiscuous}: the same body of data can be compressed by grasping very different structures. And while some compressions reflect a grasp of the relational structure of the represented domain, others reveal a grasp of the relational structure in the representational scheme itself. 

The lesson is that compression alone underdetermines what is understood. We should therefore not ask simply whether compression reflects comprehension. On the CPC account, compression always reflects some form of sensitivity to structure. The question is: the structure \emph{of what}? Is the compression facilitated by a grasp of structure in the target domain, or by a model of some more superficial structure or regularity in the way that domain is represented? Kepler's laws compress Brahe's observations by capturing regularities in planetary motion itself. The stenographer's shorthand compresses the same records by capturing regularities in their notation. Both are forms of compression, but only the former makes the heavens more intelligible.

This reframes the question of compression in LLMs as well. The issue is not whether such systems ``merely compress'' or instead compress in a way that reflects comprehension. The issue is what kind of understanding their compression reflects: are they latching onto the relational structure of the reality projected onto text, or only onto shallower relations among the words in the corpus?

These examples make it tempting to conclude that all interesting forms of compression involve grasping the structure of reality itself, whereas uninteresting forms of compression involve grasping structure in our representations of reality. But this would be too quick. Paul Humphreys \parencite*{humphreys1993} offers an instructive counterexample: the compression of logical operators using the Sheffer stroke ($\uparrow$), interpreted as NAND (``not both''). Standard presentations of propositional logic employ a family of connectives, such as conjunction ($\land$), disjunction ($\lor$), negation ($\neg$), implication ($\to$), and the biconditional ($\leftrightarrow$). Sheffer \parencite*{sheffer1913} showed that all these operators can be defined using only the Sheffer stroke. Building on this insight, Jean Nicod \parencite*{nicod1917reduction} demonstrated that the entire system of \emph{Principia Mathematica} could be derived from a single string:
\[
(A \uparrow (B \uparrow C)) \uparrow \bigl((D \uparrow (D \uparrow D)) \uparrow ((E \uparrow B) \uparrow ((A \uparrow E) \uparrow (A \uparrow E)))\bigr)
\]
Yet Nicod's axiom is far more opaque than a less compressed axiomatization for ordinary reasoning. The Sheffer stroke minimizes $L(M)$ by reducing the dictionary of logical operators to one. But this comes at the cost of increasing $L(D|M)$: ordinary instances of reasoning become harder to express and manipulate. A basic disjunction $A \lor B$ must be rebuilt by manufacturing negations and then re-negating a conjunction, yielding $(A \uparrow A) \uparrow (B \uparrow B)$. The bloat is even clearer with the biconditional, where $A \leftrightarrow B$ must be expressed via $\uparrow$-definitions of implication plus a $\uparrow$-built conjunction:
\[
\bigl((A \uparrow (B \uparrow B)) \uparrow (B \uparrow (A \uparrow A))\bigr) \uparrow \bigl((A \uparrow (B \uparrow B)) \uparrow (B \uparrow (A \uparrow A))\bigr)
\]
Since the CPC framework evaluates $L(M) + L(D|M)$, it correctly predicts that for the purposes of practical reasoning, Sheffer-stroke compression is a case of false economy: the simplicity of the primitive vocabulary is outweighed by the complexity of its application. The example also underlines the importance of the practical pressures of \textit{storability-cum-manipulability} and \textit{demonstrability}. The standard connectives are valuable precisely because they allow for efficient mental manipulation and communication.

At the same time, the Sheffer-stroke compression does reflect a nontrivial insight into the structure of logic. The example thereby points to the fact that the epistemic significance of a compression depends on \emph{what the target domain} and \emph{target task family} is. If one's purpose is not to reason about the world, but to understand the deep structure of logic itself, then grasping that all the connectives can be defined in terms of the Sheffer stroke and all the axioms can be reduced to Nicod's axiom \emph{is} a real advance in understanding. In this metalogical context, the ``data'' is the logical system itself, and it is a valuable metalogical insight that the Sheffer stroke enables a compression of that system's deep structure. Finding such a compression required real metalogical understanding.

This shows that compression at the level of a representational scheme is not always shallow. Grasping patterns in our representational apparatus can itself be a deep form of understanding. The same lesson carries back to LLMs. Even if much of their compression is achieved by modeling textual structure rather than worldly structure, grasp of textual structure can be more or less deep. An n-gram model captures superficial distributional regularities in a corpus. A system that captures English grammar models a far deeper and more valuable structure in the text, even if that structure is still linguistic rather than worldly. The CPC framework therefore takes us beyond the crude choice between ``mere compression'' and ``real understanding.'' It asks what structure is being compressed, at what depth, and for which task family.

\subsection{Comprehension Without Compression?}\label{understanding-that-resists-compression}

A complementary worry runs in the opposite direction. An influential philosophical tradition holds that the deepest forms of understanding are precisely those that \emph{resist} compression into statable principles. Aristotle's \emph{phronimos} exercises practical wisdom in particular cases without applying explicit rules; and a long line of work in epistemology and the philosophy of mind has emphasized the need for tacit and embodied forms of understanding that resist articulation. If understanding can be irreducibly implicit---the kind of thing possessed by a master craftsman, a wise judge, or a seasoned chicken sexer---then the CPC framework, with its emphasis on compressed models that can be demonstrated and transmitted, seems to leave important cases of understanding out of the picture.

\textit{Reply:} The objection helps clarify several features of the framework. First, the framework's mental models need not be explicitly articulable. The notion of a ``mental model''---an internal representation that exhibits a similar relational structure to the process it represents---admits of explicit and implicit modes. A model may be schematic, propositional, or equational; but it may also be tacit and embodied: a pattern or \emph{habitus} in one's perceptual, emotional, and inferential dispositions. A chess grandmaster's positional sense involves a structure-sensitive mental model whose presence is detectable in performance even when its contents resist propositional articulation. To grasp connections is not necessarily to be able to articulate them.

Second, compression occurs in those tacit cases, too. The \emph{phronimos} does not store every detail of every moral predicament he ever encountered. He forms compressed representations that let him recognize the salient features of a new case, anticipate likely outcomes, and respond appropriately. The fact that the understanding of the \emph{phronimos} transfers to novel terrain is itself evidence that it involves compression.

Third, ``prediction'' in the framework's sense is broader than the prediction of overt behavior or explicit data. To be predictive in the framework's sense is to reduce surprise at how a domain unfolds---including surprise about which considerations matter, which features are relevant, and which interventions will work. The phronimos's expectations are not akin to a meteorologist's forecast. They are anticipations of what a situation \emph{calls for}: which features are morally salient, which interventions will defuse the conflict or restore the relationship, which course of action will conduce to flourishing. These are structure-sensitive predictions in a perfectly recognizable sense, even though they are deployed in the service of action rather than mere description.

Fourth, the framework is graded. We do not claim that all understanding takes the elegantly principled form of Newton's laws. We claim that human understanding is \emph{under pressure} toward such forms because of the four pressures the framework identifies. Where those pressures bear most heavily on human cognition, understanding tends toward elegantly principled simplicity. Where the pressures are weaker, or where domains resist principled compression, understanding takes less principled and less compressed forms. \emph{Phronesis} sits closer to the latter pole; mathematical physics closer to the former. The framework predicts and explains this contrast rather than ruling out one end of it.

This also illuminates an instructive tension at the heart of philosophy's own treatment of \emph{phronesis}. Aristotle was clear that practical wisdom resists codification, yet ethics as a discipline has spent centuries searching for principles that compress wisdom into statable maxims. The CPC framework accommodates both sides of this tension. The cognitive state of moral understanding is partly tacit, but the social uses of moral concepts---the need to teach, assess, coordinate, and justify---exert constant pressure toward articulation. The framework does not adjudicate this tension; it explains why the tension is unavoidable for creatures who must rely on and learn from one another (after all, we presumably owe the very existence of Aristotle's lecture notes on \emph{phronesis} to the pressure to teach).

\section{Conclusion}\label{conclusion}
Past performance, the saying goes, is the best predictor of future performance. We have argued that we have evolved a better strategy: probing for understanding. The concept arose as an efficient proxy for robust competence, enabling us to identify whom to trust and whom to learn from. From those social functions, four pressures arise—predictivity, storability-cum-manipulability, demonstrability, and transmissibility—whose convergence yields the CPC framework: to understand a domain is to possess a mental model that grounds robust competence by making the domain less surprising, and what becomes less surprising can be more economically described. Compression is not identical with comprehension; it is its representational shadow.

This framework casts the quest for machine understanding in a new light. It allows us to see the preoccupation with whether AI models ``generalize'' to new data as the latest instantiation of a longstanding human preoccupation with robust competence. Yet one lesson of the CPC framework is that understanding cannot be read off from predictive success alone. The critical question is not simply \emph{whether} a system compresses or generalizes beyond its training data, but \emph{how} it does so---in particular, \emph{what structure} its compression reflects: is the system latching onto shallow regularities in the corpus, deeper linguistic structure, or the relational structure of the target domain?

Another lesson of the framework, finally, is that where predictive accuracy is the dominant pressure, we should expect not elegant understanding, but opaque understanding: high predictivity coupled with incommunicable complexity, of the kind we find in some of the most powerful predictive systems—from large-scale climate models to today's neural networks. This is the expected shape of competence freed from the obligations to demonstrate and transmit understanding. The human ideal of understanding, by contrast, is not simply a matter of making the world less surprising. It is a socially disciplined form of predictive compression: one that can be displayed, challenged, and taught. Understanding, in the form we have come to prize, is not a feature of minds simpliciter, but of minds that must answer to other minds.

 \printbibliography[title={List of Works Cited}]
\end{document}